\documentclass[final]{cvpr}
\usepackage{nopageno}
\usepackage{times}
\usepackage{epsfig}
\usepackage{graphicx}
\usepackage{amsmath}
\usepackage{amssymb}
\usepackage{dsfont}
\usepackage{bm}
\usepackage{enumitem}
\usepackage[utf8]{inputenc}
\DeclareUnicodeCharacter{2212}{-}

\usepackage[pagebackref=true,breaklinks=true,colorlinks,bookmarks=false]{hyperref}

\def\cvprPaperID{2873} % *** Enter the CVPR Paper ID here
\def\confYear{CVPR 2021}

\begin{document}

%%%%%%%%% TITLE
\title{Coarse-to-Fine Domain Adaptive Semantic Segmentation with Photometric Alignment and Category-Center Regularization}

\author{Haoyu Ma$^{1}$\thanks{These authors have equal contribution.}\quad\quad Xiangru Lin$^{1}$\footnotemark[1]\quad\quad Zifeng Wu$^{2}$ \quad\quad Yizhou Yu$^{1}$\thanks{Corresponding author} \vspace{2mm}\\
$^1$The University of Hong Kong   \quad\quad\quad $^2$Deepwise AI Lab\\
{\tt\footnotesize mahaoyu@connect.hku.hk},{\tt\footnotesize xrlin2@cs.hku.hk}, {\tt\small wuzifeng@deepwise.com}, {\tt\small yizhouy@acm.org}
 \vspace{-0mm}
}

\maketitle

%%%%%%%%% ABSTRACT
\begin{abstract}
Unsupervised domain adaptation (UDA) in semantic segmentation is a fundamental yet promising task relieving the need for laborious annotation works. However, the domain shifts/discrepancies problem in this task compromise the final segmentation performance. Based on our observation, the main causes of the domain shifts are differences in imaging conditions, called image-level domain shifts, and differences in object category configurations called category-level domain shifts. In this paper, we propose a novel UDA pipeline that unifies image-level alignment and category-level feature distribution regularization in a coarse-to-fine manner. Specifically, on the coarse side, we propose a photometric alignment module that aligns an image in the source domain with a reference image from the target domain using a set of image-level operators; on the fine side, we propose a category-oriented triplet loss that imposes a soft constraint to regularize category centers in the source domain and a self-supervised consistency regularization method in the target domain. Experimental results show that our proposed pipeline improves the generalization capability of the final segmentation model and significantly outperforms all previous state-of-the-arts.
\end{abstract}

%%%%%%%%% BODY TEXT
\section{Introduction}
% introduce the background
Semantic segmentation is a fundamental computer vision task that aims to assign a semantic category label to every pixel in an image. It has been widely used in many important downstream tasks such as autonomous driving~\cite{autodrive1,autodrive2} and medical image analysis~\cite{medical1,medical2,medical3}. Recent state-of-the-art methods on semantic segmentation are primarily deep learning based ~\cite{deeplab,oldfcn,crf} and require a large number of high quality annotated ground-truth data which are difficult to obtain especially in practical applications. Unsupervised domain adaptation semantic segmentation is an alternative method to solve the data scarcity problem where it generalizes models trained on the source domain composed of synthetic images and labels to perform well on the target domain composed of real world images only~\cite{gta5, uda1, stuff_things,sda1,sda2}. However, the problem is that semantic segmentation models trained merely on synthetic data exhibit poor performance on real world images due to the differences in multiple aspects (also called domain shifts/discrepancies), including exposure, contrast, lighting, object shape and surface textures, between the source domain and the target domain. Therefore, matching the distributions between the source and target domains to learn domain-invariant representations is crucial to solve the domain shifts.

% analyze the problem and state motivation
Although the domain shifts could be caused by multiple factors, based on our observation, the primary causes can be summarized into two groups, namely image-level domain shifts and category-level domain shifts. For the image-level domain shifts, these refer to the differences in imaging conditions, such as lighting and settings in the camera imaging pipeline. Existing works on solving image-level domain shifts through image style transfer generally utilize deep models such as generative models and image-to-image translation models~\cite{cgan, cgan2} while another line of research focuses on using Fourier transformation~\cite{FDA}. These methods have proven that transferring image style of one domain to another domain can bring the two domains closer. However, the downside of these methods is that they either require to carry out a computationally expensive training process for the deep models or generate inferior style-transferred output images as shown in Figure ~\ref{fig:qualitative_gpa}. 

Despite the fact that the domain gap can be minimized by global alignment methods such as the above, there is no guarantee that samples from different object categories in the target domain can be well separated. This is because some categories are naturally close to others in terms of body shape, pose and textures. To solve this problem, existing methods adopt category anchors computed on the source domain to guide the alignment between the two domains~\cite{CAG,stuff_things}, which can be regarded as a hard constraint on the category centers. The problem of this design is that it does not regularize the distance between different category features, and categories with similar feature distributions in the source domain also have similar distributions in the target domain, which results in erroneous classification results especially when no supervision information is available in the target domain. Our experimental results have demonstrated that imposing soft regularization methods on category distributions can improve the model's capacity to adjust the relative magnitude of inter-category and intra-category feature distances. 
%the features from different categories often distribute unevenly and some categories are quite close to each other, which will result in erroneous classification results especially when no supervision information is available in the target domain. Our experiment results have demonstrated that such fixed category centers inevitably limits the model's capacity to adjust the relative magnitude of inter-category and intra-category feature distances. 

According to the analysis above, performing alignment from either image-level perspective or category-level perspective alone will not solve the domain shifts reasonably. Therefore, we approach the problem from a different perspective and propose a novel and efficient pipeline that unifies image-level alignment and category-level feature distribution regularization in a coarse-to-fine manner. In general, on the coarse side, we propose a novel and efficient image-level alignment module to coarsely align the two domains; on the fine side, we introduce a new category-oriented triplet loss to softly regularize the category centers in the source domain and propose a self-supervised consistency regularization method in the target domain. By addressing both level of domain shifts simultaneously, we can significantly improve the performance of our proposed domain adaptation method.

\textbf{Coarse Alignment.} To solve the image-level domain shifts discussed above, we propose a global photometric alignment (GPA) module that aligns an image in the source domain with a reference image from the target domain using a set of image-level operators. %Our method differs from other generative methods and fourier transformation based methods in two aspects
Our method is superior to other generative methods and Fourier transformation based methods in two aspects: first, compared to the generative counterparts, our method requires no extra training process and produces stochastic image results; second, the quality of the translated image and the performance of our method is comparable to its generative counterpart and is superior to that of Fourier transformation based methods.

\textbf{Category-level Feature Distribution Regularization.} To address category-level domain shifts on the fine side, in addition to the common strategy of using pseudo labels for the target domain, we propose two novel regularization methods for the source and target domains respectively. First, considering the fact that there are annotated ground truth labels in the source domain, we propose a category-oriented triplet loss (CTL) that imposes a soft constraint to regularize category centers calculated using the source image pixel features, which actively enlarges the distances among category centers, making inter-category distances in a high-level feature space larger than intra-category distances by a predefined margin. Second, inspired by the commonly used self-supervised learning methods: consistency regularization and pseudo-labeling, we propose a simple yet effective consistency regularizer for the target domain, called target domain consistency regularization (TCR), which constrains the prediction on an augmented target image to be consistent with the pseudo label of the corresponding non-augmented image, forcing the class labels of similar semantic contents to be consistent in the target domain.

% our contribution and summary
In conclusion, this paper has the following contributions:
\begin{itemize}[noitemsep, nolistsep]
  \item We propose a novel coarse-to-fine domain adaptive semantic segmentation pipeline that seamlessly combines coarse image-level alignment with finer category-level feature distribution regularization. 
  %Specifically, the source images are first aligned with the target images photometrically by our GPA module to perform coarse alignment online, and then category-level feature distributions are further adjusted by our proposed category-level regularization methods.
  \item We introduce two novel and effective category-level regularization methods for the source and target domains respectively. The first one is called category-oriented triplet loss that regularizes category centers in the source domain while the second one performs target domain consistency regularization.
  \item Our method outperforms all previous methods, achieving new state-of-the-art performance on both GTA5$\rightarrow$Cityscapes and SYNTHIA$\rightarrow$Cityscapes benchmarks.
\end{itemize}

%------------------------------------------------------------------------
\section{Related Work}
%Deep learning has become very popular because of their powerful capability to extract features in the research area of computer vision~\cite{}. However, the strong capability of extract features also make deep learning models prune to over-fitting, and the performance of the model is largely depends on the attribute of the datasets. As a result, the topic of how to generalize the model capability to overcome domain shift has been widely studied in areas such as super-resolution~\cite{}, classification~\cite{}, segmentation~\cite{}. In our study, we focus on the unsupervised semantic segmentation problems.
Since our proposed domain adaptation pipeline is mostly related to photometric alignment based~\cite{da1, da2,FDA} and category-based domain adaptation methods~\cite{stuff_things,CAG}, we focus on these two types of work in this section.

\textbf{Photometric Alignment.} Previous works on domain adaptation~\cite{BDL,dagan1,dagan2,dagan3,dagan4,take_a_look,stuff_things,dagan5} have applied adversarial models, such as GAN~\cite{gan,cgan} and CycleGAN~\cite{cgan2}, to achieve photometric alignment results. Adversarial training makes a model capable of transferring image styles from one domain to another to significantly reduce the photometric differences between the two domains in the original image space~\cite{dagan3,stuff_things,BDL}. Then a segmentation model trained on (style transferred) source domain images can be applied to target domain images~\cite{BDL,stuff_things}. However, adversarial models are hard to train. Many researchers~\cite{fcns, CAG, stuff_things} have also shown that models based on adversarial training generally align distributions from different domains, but do not actually obtain mappings between features from different domains.
%domain adaptation for segmentation does not require artistic effect which is commonly seen in style transfer tasks~\cite{}. And the physical meanings in domain shift is also more straightforward: they are usually because of the different settings of the media collection devices, rather than the intervention of people, and we believe global domain shift can be tackled in the image-level by basic image processing techniques.
Other types of photometric alignment methods for unsupervised semantic segmentation are rare. One method was proposed in \cite{FDA} to align the source and target domains by simply replacing the low frequency component in a source domain image with its counterpart in the target domain reference image. However, such simple substitution of frequency components leaves unsatisfactory visual artifacts, and the performance of the model trained on the aligned samples relies heavily on a multi-band ensemble. On the contrary, our method is different from previous method in that it has a light-weight photometric alignment strategy which does not require to carry out a computationally expensive training process and more importantly, produces comparable (superior) performance and image quality with respect to its generative (Fourier transformation based) counterpart.

\textbf{Category-Based Methods.}  Category labels/predictions were introduced in \cite{fcns,take_a_look,dagan1,fgan} to enforce global semantic constraints on the distribution of predicted labels. The proposed methods in \cite{CAG} and \cite{stuff_things} take one step further. They map penultimate target domain image features, that are used for generating pseudo labels in the output layer, to the corresponding features of the source domain image. However, in their work, category feature centroids~\cite{CAG} or instance features~\cite{stuff_things} in the source domain serve as anchors for category-based feature alignment, which does not explicitly enlarge the margins between the centers. This alignment strategy can be problematic because category anchors close in the source domain are likely hard to separate in the target domain as well. Our work differs from theirs in the following aspects: first, we propose a category-oriented triplet loss for the source domain that imposes a soft constraint to regularize category centers, actively making inter-category distances in a high-level feature space larger than intra-category distances by a specified margin; second, to further constrain category-level feature distributions in the target domain, we force the predictions on augmented target domain images to be consistent with the pseudo labels, generated by the segmentation model, of the corresponding non-augmented images, which is a self-supervision based consistency regularization method.

\section{Method}
\subsection{Coarse-to-Fine Pipeline}
The key idea underlying our domain adaptation pipeline is intuitive: first, we exploit the photometric differences in the two domains and coarsely align the source domain images with the target domain images to minimize the domain shift; then, we regularize category-level feature distributions by setting constraints on inter-class center distances and intra-class feature variations.

\begin{figure*}[ht]
	\centering
	\includegraphics[width=0.8\linewidth]{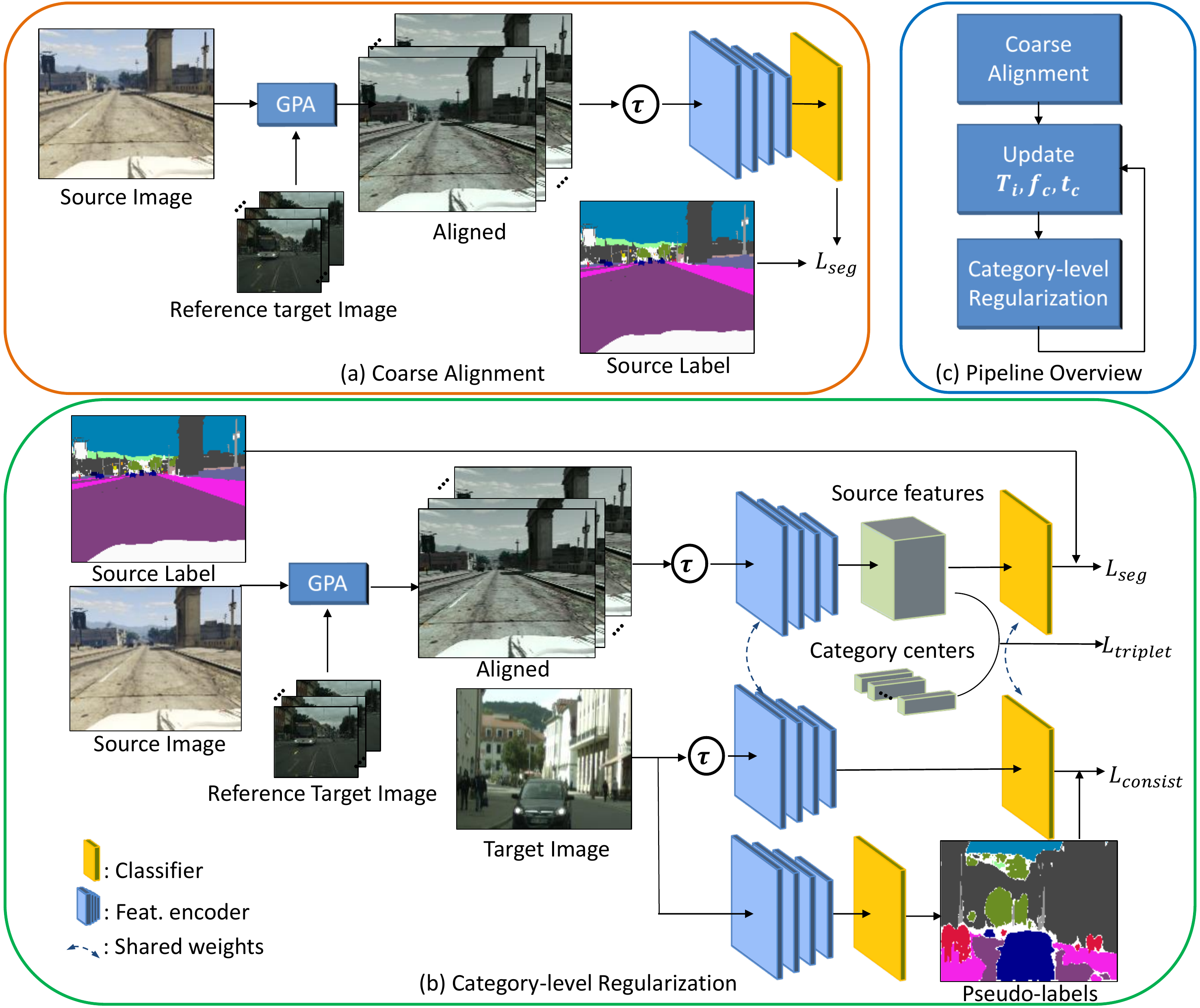} % Reduce the figure size so that it is slightly narrower than the column.
	\caption{(a) First, the global photometric alignment (GPA) module is used to coarsely align the source and target domain images to train the initialized segmentation model $T_{0}$. (b) Then, we train the category-level feature distribution regularization step with the calculated category feature center $f_{c}$ and pseudo-label threshold $t_{c}$ for each category $c$. The category-oriented triplet loss is applied to the source domain and the consistency regularization is used in the target domain to jointly regularize the category-level feature distribution. (c) The overall pipeline is trained in an iterative self-supervised manner with $T_{i}$, $f_{c}$, and $t_{c}$ updated at each step $i$.}
	\label{fig:pipeline}
\end{figure*}

\textbf{Step 0: Coarse Alignment.} Define $\bm{{\rm M}}=\{ m_k \}_{k=1}^{N_s}$ as the source domain training set, where $m_k$ is a source domain image and $N_s$ is the number of images in the source domain training set. Similarly, the target domain training set is defined as $\bm{{\rm N}}=\{ n_k \}_{k=1}^{N_t}$. Our proposed GPA module converts a source domain image $m$ in the training batch and a randomly selected target domain reference image $n$ into Lab color space as $(L_m, a_m, b_m)$ and $(L_n, a_n, b_n)$. Then the histogram mapping function $f_{match}$ is applied to $a_m$ and $b_m$, and gamma correction function $f_{gamma}$ is applied to $L_m$ to form $(f_{gamma}(L_m), f_{match}(a_m), f_{match}(b_m))$. The image is then converted to RGB space as aligned image $m'$ to construct aligned source domain training set \bm{${\rm M'}$}. Then, a stochastic function $\tau$ is applied to produce an augmented version of every image in \bm{${\rm M'}$}. A segmentation model $T_0$ is trained based on all style-transferred images $\tau(\bm{{\rm M}'})$ with segmentation loss $L_{seg}$. 

\textbf{Step 1: Category-level Feature Distribution Regularization.} In this step, we train a segmentation model $T_1$ with $\tau(\bm{{\rm M}'})$ and $\bm{{\rm N}}$. We apply the segmentation model $T_0$ to all images in the target domain to produce a feature vector and a class probability vector at every pixel. The category corresponding to the largest value of the probability vector is defined to be the pseudo label at the pixel, and the largest probability value itself defines the confidence of the pseudo label. We further pre-define the pair of probability threshold $P_{h}$ and percentage threshold $p$ for all categories. The latter gives rise to a category specific probability threshold $P_{s, c}$, meaning $p\%$ pixels in the category have confidence above $P_{s, c}$. Thus the final confidence threshold for category $c$ is $t_c=\min(P_{h}, P_{s, c})$, and any pseudo labels in this category with a confidence higher than $t_c$ are considered valid and added to the segmentation loss $L_{seg}$. The remaining pixels are left out during backpropagation. Then category center $f_c$ for every category $c$ are also calculated as the L2 normalized mean of all pixel features with category $c$ as the ground truth label in the source domain. In addition to aligned training set $\bm{{\rm M}'}$ and cross-entropy loss $L_{seg}$, we impose a category-oriented triplet loss $L_{triplet}$ on the segmentation model $T_1$ in the source domain to enlarge inter-category distances, and a target domain consistency loss $L_{consist}$ to regularize category-level feature distributions in the target domain. Then we finetune model $T_0$ $U$ iterations to produce the model $T_1$ by minimizing $L_{seg}+L_{triplet}+L_{consist}$.

\textbf{Step 2 to K: Iterative Self-Supervised Training.} Model $T_1$ trained in Step 1 can be further improved with iterative steps similar to Step 1. Such an iterative approach is frequently called self-supervised training in the area of unsupervised domain adaptation for semantic segmentation~\cite{BDL,CAG,FDA,stuff_things}. The same Step 1 is executed except that model $T_{i-1}$ instead of $T_0$ is used as the pretrained model to generate pseudo-labels and category centers $f_c$. This process is repeated for $K-1$ times. The overall pipeline of our proposed coarse-to-fine method is shown in Figure~\ref{fig:pipeline}. 

\subsection{Global Photometric Alignment}
Since the global domain shift mostly affects low-level pixel attributes, which are irrelevant to pixel-wise category labels, we propose global photometric alignment (GPA) to align images from the source with images from the target domain. We observe that the spatial lightness distribution of an image can be very complicated under certain circumstances while the spatial color distribution of $a$ and $b$ have similar bell-shaped histograms. Therefore, we treat lightness and color differently and perform classic histogram matching~\cite{DIP} between the source domain image and the target domain reference image only on color channels $a$ and $b$ to avoid introducing artifacts commonly seen in histogram matching results.

\begin{figure}[ht]
	\centering
	\includegraphics[width=0.95\linewidth]{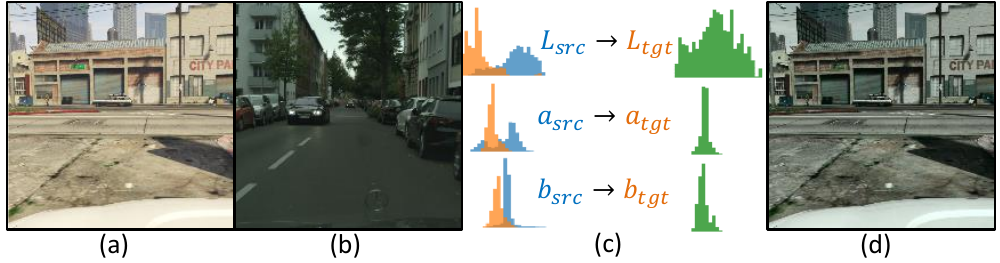} 
	\caption{(a)Input source domain image and (b) a randomly chosen target domain image is aligned in (c) Lab channels to generate (d) aligned image.}
	\label{fig:aug}
\end{figure}

\textbf{Lightness Gamma Correction.} $L$ channel, on the other hand, is much more diversified among images. This is because light interacts with the 3D structure of a scene in a complicated manner. Simple histogram matching function gives rise to large areas of overexposure and fake structures. Thus, instead of strictly enforcing the mapping constraint prescribed by histogram matching for every histogram bin, we choose to only constrain the mean value of the lightness channel in the source domain image and make it equal to the mean value of the target domain reference image. Here, we choose the power-law function. The difference between our proposed method and the classic gamma correction is that our function coefficients are automatically calculated with given source-target image pairs rather than user-defined. Specifically, we define $f_{gamma}(L)=L^{\gamma}$, where $L$ is the normalized lightness value.
Then the mean value constraint can be written as
\begin{equation}
\sum_{L}Lp^m_s(f_{gamma}(L)) = \sum_{L}Lp^m_s(L^{\gamma}) = \sum_{L}Lp^n_t(L)
\end{equation}
, where $p^m_s$ is the lightness histogram of source image $m$, and $p^n_t$ is the lightness histogram of target reference image $n$. This is a nonlinear equation and $\gamma$ can be solved numerically. $\gamma=1$ when it is an identical transformation.
In practice, to prevent $\gamma$ from deviating too much away from 1, we introduce a regularization term into the following minimization,
\begin{equation}
\label{equ:histgamma}
\gamma* = \arg \min_\gamma \left( \sum_{L}Lp^m_s(L^{\gamma}) - \sum_{L}Lp^n_t(L) \right)^2+\beta(\gamma-1)^2,
\end{equation}
%$$f_{m,n}(x) = 255\cdot\left( \frac{x}{255} \right)^{\gamma*}$$
which is a simple convex optimization problem with only one variable $\gamma$, and can be easily solved with few steps of gradient descent. The process of proposed GPA module is illustrated in Figure~\ref{fig:aug}.

%\begin{algorithm}
%	\caption{Global Chromatic Alignment}
%	\label{alg:GCA}
%	\begin{algorithmic}[1]
%		\Require Source image $src$, Reference target image $ref$
%		
%		\Function {Align}{$src, ref$}
%		\State $L_{src}, a_{src}, b_{src} \gets RGB2Lab(src)$
%		\State $L_{tgt}, a_{tgt}, b_{tgt} \gets RGB2Lab(tgt)$
%		\State $p_{src} \gets calculateHistogram(L_{src})$
%		\State $p_{tgt} \gets calculateHistogram(L_{tgt})$
%		\State $\gamma*\gets\arg \min_\gamma \left( \sum_{l}lp^m_s(l^{\gamma}) - \sum_{l}lp^n_t(l) \right)^2$
%		\State $L_{result} \gets 255\cdot\left( \frac{x}{255} \right)^{\gamma*}$
%		\State $a_{result} \gets matchHistogram(a_{src}, a_{tgt})$
%		\State $b_{result} \gets matchHistogram(b_{src}, b_{tgt})$
%		\State $result \gets Lab2RGB(L_{result}, a_{result}, b_{result}) $
%		
%		\Return{$result$}
%		\EndFunction
%	\end{algorithmic}
%\end{algorithm}
%
%After applying operations to L, a, b channels, we consider the image is roughly aligned in the low level. The summarized steps for proposed Global Chromatic Alignment are illustrated in Algorithm~\ref{alg:GCA}.
\subsection{Training Loss}
The only training loss during coarse alignment step is the segmentation cross-entropy loss. The overall loss function we use during the category-level stage consists of three parts: the cross-entropy classification loss, a category-oriented triplet loss, and a target domain consistency regularization loss. 
%Based on our observation, the cross-entropy domain already minimize the features in the same category, and the features are getting close to the category center across self-supervision stages. But the self-supervised training for the source domain dominates the training, and the distances between different category centers are quite different: some margins are relatively high while others are low. Therefore, we also introduce the category-oriented triplet loss for this specific task. \\
%We adopt ResNet101 and ASPP as our segmentation model. The source domain data is processed with global chromatic alignment and enriched with categorical data augmentation. Then model is feed forward to the segmentation model. In order to improve the generality of the model, [CAG] proposed an "anchor" loss in addition to the conventional classification losses for source labels and target pseudo-labels.

\textbf{Category-oriented Triplet Loss.} Even though the features learned with the GPA module are domain-invariant to some extent, the cross entropy losses used in previous training does not explicitly control the category-wise feature distribution. Therefore, the model learned with the GPA module using cross-entropy losses is coarsely aligned. Pixel features are distributed unevenly among different categories and some category centers are close to each other. To tackle this issue, we propose a category-oriented triplet loss that aims to further push the category-wise features closer to the corresponding category centers and further from other category centers. 
%This loss is complementary to cross-entropy loss and is applied to source domain images only. The proposed novel category-oriented triplet loss (CTL) not only exploits hard training samples in the source domain but also controls the inter-category feature distribution, which differentiates the pixel features of each category from the rest of the categories.
Let $x_{i,j}$ be the pixel-wise features in the feature map of the second last layer, and $y_{i,j}$ be the ground truth pixel-wise labels of a source domain image. The category center of category $c$ is calculated as follows,

\begin{equation}
f_c = G(\frac{1}{N_c} \sum_s \sum_i \sum_j  1\left( y_{i,j}=c \right)x_{i,j})
\end{equation}

where $N_c$ is the total number of pixels in category $c$ and $s$ is the source domain image index, and $G$ is a L2 normalization function. Note that this L2 normalization is crucial to keep the category centers on the unit sphere to avoid scaling issue. The centers are updated after the training and this allows the centers become further and further from each other on the sphere surface.

%We find this center loss an effective compensation for the conventional cross-entropy and "smooth" the intra-category variance during training, thus we keep this loss in our experiment.
%The category centers or pseudo-labels are calculated with pretrained models as in many literatures regarding of the unsupervised semantic segmentation [CAG, FDA, Intra-domain, Differential].
Our category-oriented triplet loss is formulated as follows,
\begin{equation}
\label{equ:triplet}
\begin{aligned}
L_{triplet} =& \frac{1}{N_{s}} \sum_s \sum_C \sum_i \sum_j  \max ( \left\| G(x_{i,j})-f_{c=C} \right\| \\ & - \left\| G(x_{i,j})-f_{c\neq C} \right\| +\alpha, 0),
\end{aligned}
\end{equation}
where $N$ is the total number of pixels in all images, and $\alpha$ is a prescribed margin. The loss would be zero if every feature $x_{i,j}$ is at least $\alpha$ closer to its own category center than other category centers. Note that we only have reliable category labels for the source domain, thus we only apply the category-oriented triplet loss to the source domain images.

\begin{figure}[ht]
	\centering
	\includegraphics[width=\linewidth]{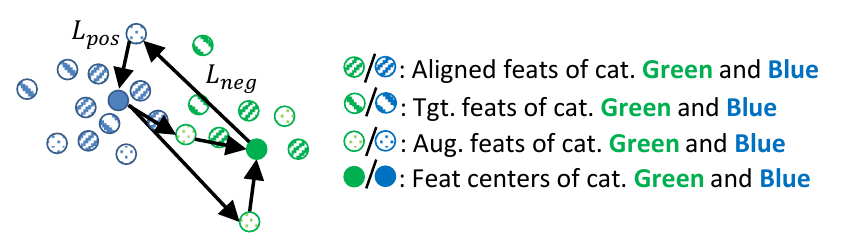} % Reduce the figure size so that it is slightly narrower than the column.
	\caption{Our proposed category-oriented triplet loss exploits hard samples and further enlarge category margins.}
	\label{fig:triplet}
\end{figure}

The working principals of our proposed category-oriented triplet loss is illustrated in Figure~\ref{fig:triplet}. In cooperation with proposed photometric alignment and data augmentation in the source domain, our proposed triplet loss exploits hard samples in the coarsely aligned source domain and further improves the generalization capability of the trained model, which serves as complementary to cross-entropy loss. 

\textbf{Target Domain Consistency Regularization.} Category-oriented triplet loss is designed to regularize category-wise features in the source domain where the annotated ground truth labels are available. However, this is not the case in the target domain where no labels are provided. Consistency regularization is an important component of many recent state-of-the-art self-supervised learning algorithms, which utilizes unlabeled data by relying on the assumption that the model should output similar predictions when fed perturbed versions of the same image~\cite{consistency, fixmatch}. Motivated by this, we propose a target domain consistency regularization method shown in Figure~\ref{fig:pipeline} to perform category-level feature distribution regularization in the target domain.

%Simple data augmentation will improve the generality of the model and has been used in previous works\cite{cgan}. However, they are only used in source domains because of the lack of labels in the target domain. But training samples of target domain is usually many times fewer than source domain, and mere data augmentation on the source domain might further make the training set more imbalanced. Comparing to the large quantity of available reliable labels in the source domain, the target domain are definitely in a more desperate need for more diversified samples. 
%Inspired by current development in the self-supervised learning works~\cite{consistency, fixmatch}, we propose a simple but efficient target domain augmentation regularization method shown in Figure~\ref{fig:pipeline}. 

% write a generalized version of fixmatch and state our setting in this task
The idea of our proposed consistency regularization is simple: given a target domain image $n_{k}$, with the trained segmentation model $T_{i-1}$, we extract a pseudo label $\hat{y}^{k}_j$  at every location $j$ by feeding $n_k$ to $T_{i-1}$ followed by applying $\arg\max(.)$; and the corresponding pixel prediction is converted to a hard label vector $\mathds{1}_{[c=\hat{y}^{k}_j]}$; then, we apply the stochastic function $\tau$ to $n_{k}$ to obtain a perturbed version $n'_{k}$; then, we feed $n'_{k}$ to $T_{i}$ to obtain a prediction ${p}'^k_{j}$ at every location $j$ in the perturbed image; finally, ${p}'^k_{j}$ is forced to be consistent with $\hat{y}^{k}_j$  by using a cross entropy loss function at pixel locations whose largest class probability is above the previously defined category-level confidence threshold $t_c$. By doing this, category-level feature distributions in the target domain are regularized under the supervision of valid pseudo labels. The overall formula is defined as follows,
\begin{equation}
\label{equ:histgamma}  
\begin{aligned}
L_{cst} &= \sum_j \bm{1}(\max(T_{i-1}(n_{k})|_j)\geq t_c)\mbox{CELoss}(\mathds{1}_{[c=\hat{y}^{k}_j]}, {p}'^k_{j}), \\
\hat{y}^k_j &= \arg\max(T_{i-1}(n_{k})|_j),\\
{p}'^k_{j} &= T_{i}(n'_{k})|_j.
\end{aligned}
\end{equation}
It is important to use trained model $T_{i-1}$ rather than model $T_i$ to generate pseudo labels. This is because $T_i$ is still being trained and unstable. Fluctuating pseudo labels generated by $T_i$ would be catastrophic to the training process. Experimental results show this consistency regularization method is very effective even though the idea is simple.

\begin{table*}[t]
	\centering
	\resizebox{0.95\linewidth}{!}{
	\begin{tabular}{l l l l l l l l l l l l l l l l l l l l l l}
		\hline
		& \rotatebox{90}{road} & \rotatebox{90}{sidewalk} & \rotatebox{90}{building} & \rotatebox{90}{wall} & \rotatebox{90}{fence} & \rotatebox{90}{pole} & \rotatebox{90}{light} & \rotatebox{90}{sign} & \rotatebox{90}{vege}. & \rotatebox{90}{terrace} & \rotatebox{90}{sky} & \rotatebox{90}{person} & \rotatebox{90}{rider} & \rotatebox{90}{car} & \rotatebox{90}{truck} & \rotatebox{90}{bus} & \rotatebox{90}{train} & \rotatebox{90}{motor} & \rotatebox{90}{bike} & mIoU \\
		\hline
		BDL~\cite{BDL} & 91.0 & 44.7 & 84.2 & 34.6 & 27.6 & 30.2 & 36.0 & 36.0 & 85.0 & \textbf{43.6} & 83.0 & 58.6 & 31.6 & 83.3 & 35.3 & 49.7 & 3.3 & 28.8 & 35.6 & 48.5 \\
		IDA~\cite{intra_domain} & 90.6 & 36.1 & 82.6 & 29.5 & 21.3 & 27.6 & 31.4 & 23.1 & 85.2 & 39.3 & 80.2 & 59.3 & 29.4 & 86.4 & 33.6 & 53.9 & 0.0 & 32.7 & 37.6 & 46.3 \\
		DTST~\cite{stuff_things} & 90.6 & 44.7 & 84.8 & 34.3 & 28.7 & 31.6 & 35.0 & 37.6 & 84.7 & 43.3 & 85.3 & 57.0 & 31.5 & 83.8 & \textbf{42.6} & 48.5 & 1.9 & 30.4 & 39.0 & 49.2\\
		FGGAN~\cite{fgan} & 91.0 & 50.6 & 86.0 & \textbf{43.4} & \textbf{29.8} & 36.8 & 43.4 & 25.0 & \textbf{86.8} & 38.3 & 87.4 & 64.0 & \textbf{38.0} & 85.2 & 31.6 & 46.1 & 6.5 & 25.4 & 37.1 & 50.1\\
		FDA~\cite{FDA} & \textbf{92.5} & 53.3 & 82.3 & 26.5 & 27.6 & 36.4 & 40.5 & 38.8 & 82.2 & 39.8 & 78.0 & 62.6 & 34.4 & 84.9 & 34.1 & \textbf{53.1} & 16.8 & 27.7 & 46.4 & 50.4 \\
		CAG~\cite{CAG} & 90.4 & 51.6 & 83.8 & 34.2 & 27.8 & \textbf{38.4} & 25.3 & \textbf{48.4} & 85.4 & 38.2 & 78.1 & 58.6 & 34.6 & 84.7 & 21.9 & 42.7 & 41.1 & 29.3 & 37.2 & 50.2 \\
		\hline
		coarse align. (ours) & 83.9 & 37.5 & 82.7 & 28.7 & 18.9 & 35.3 & 41.3 & 31.1 & 85.2 & 29.5 & 86.6 & 62.8 & 30.9 & 82.4 & 23.0 & 39.3 & 33.0 & 26.0 & 39.7 & 47.3\\
		coarse-to-fine (ours) & \textbf{92.5} & \textbf{58.3} & \textbf{86.5} & 27.4 & 28.8 & 38.1 & \textbf{46.7} & 42.5 & 85.4 & 38.4 & \textbf{91.8} & \textbf{66.4} & 37.0 & \textbf{87.8} & 40.7 & 52.4 & \textbf{44.6} & \textbf{41.7} & \textbf{59.0} & \textbf{56.1}\\
		\hline
	\end{tabular}}
    \vspace{1mm}
	\caption{Performance comparison with state-of-the-art methods on the GTA5$\rightarrow$Cityscapes task. Results after only coarse alignment and whole coarse-to-fine pipeline are both presented.}
	\label{tab:gta2city}
\end{table*}

\begin{figure*}[ht]
	\centering
	\includegraphics[width=0.8\linewidth]{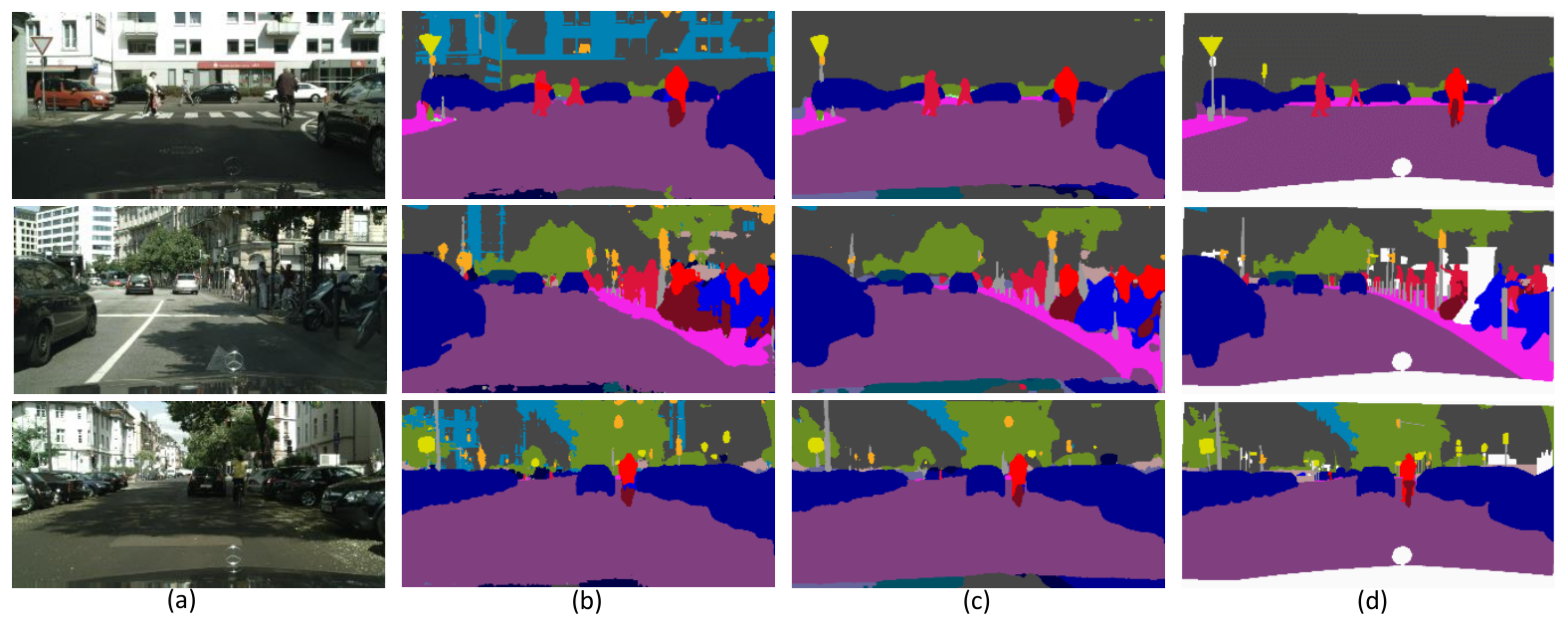} % Reduce the figure size so that it is slightly narrower than the column.
	\caption{Qualitative analysis on GTA5$\rightarrow$Cityscapes task. (a) Input images, (b) CAG~\cite{CAG}, (c)Ours, (d) Labels}
	\label{fig:qualitative}
\end{figure*}

\begin{table*}[t]
	\centering
	\resizebox{0.95\linewidth}{!}{
	\begin{tabular}{l l l l l l l l l l l l l l l l l l l l l l}
		\hline
		& \rotatebox{90}{road} & \rotatebox{90}{sidewalk} & \rotatebox{90}{building} & \rotatebox{90}{wall} & \rotatebox{90}{fence} & \rotatebox{90}{pole} & \rotatebox{90}{light} & \rotatebox{90}{sign} & \rotatebox{90}{vege}. & \rotatebox{90}{sky} & \rotatebox{90}{person} & \rotatebox{90}{rider} & \rotatebox{90}{car} & \rotatebox{90}{bus} & \rotatebox{90}{motor} & \rotatebox{90}{bike} & mIoU & mIoU*\\
		\hline
		BDL~\cite{BDL} & \textbf{86.0 } & \textbf{46.7 } & 80.3 & - & - & - & 14.1 & 11.6 & 79.2 & 81.3 & 54.1 & 27.9 & 73.7 & 42.2 & 25.7 & 45.3 & - & 51.4 \\
		IDA~\cite{intra_domain} & 84.3 & 37.7 & 79.5 & 5.3 & 0.4 & 24.9 & 9.2 & 8.4 & 80.0 & 84.1 & 57.2 & 23.0 & 78.0 & 38.1 & 20.3 & 36.5 & 41.7 & 48.9 \\
		DTST~\cite{stuff_things} & 83.0 & 44.0 & 80.3 & - & - & - & 17.1 & 15.8 & 80.5 & 81.8 & 59.9 & 33.1 & 70.2 & 37.3 & 28.5 & 45.8 & - & 52.1 \\
		FGGAN~\cite{fgan} & 84.5 & 40.1 & 83.1 & 4.8 & 0.0 & 34.3 & \textbf{20.1 } & 27.2 & 84.8 & 84.0 & 53.5 & 22.6 & \textbf{85.4 } & \textbf{43.7 } & 26.8 & 27.8 & 45.2 & 52.5 \\
		FDA~\cite{FDA}  & 79.3 & 35.0 & 73.2 & - & - & - & 19.9 & 24.0 & 61.7 & 82.6 & 61.4 & 31.1 & 83.9 & 40.8 & \textbf{38.4 } & 51.1 & - & 52.5 \\
		CAG (13 classes)~\cite{CAG}  & 84.8 & 41.7 & \textbf{85.5 } & - & - & - & 13.7 & 23.0 & \textbf{86.5 } & 78.1 & \textbf{66.3 } & 28.1 & 81.8 & 21.8 & 22.9 & 49.0 & - & 52.6 \\
		CAG (16 classes)~\cite{CAG} & 84.7 & 40.8 & 81.7 & 7.8 & 0.0 & 35.1 & 13.3 & 22.7 & 84.5 & 77.6 & 64.2 & 27.8 & 80.9 & 19.7 & 22.7 & 48.3 & 44.5 & - \\ 
		\hline
		coarse align. (ours) & 64.0 & 25.7 & 73.9 & 9.6 & 0.8 & 33.3 & 12.3 & 25.9 & 81.6 & 85.5 & 62.4 & 26.2 & 80.6 & 30.9 & 26.8 & 23.8 & 41.5 & 47.7 \\
		coarse-to-fine (ours) & 75.7 & 30.0 & 81.9 & \textbf{11.5} & \textbf{2.5} & \textbf{35.3} & 18.0 & \textbf{32.7} & 86.2 & \textbf{90.1} & 65.1 & \textbf{33.2} & 83.3 & 36.5 & 35.3 & \textbf{54.3} & \textbf{48.2 } & \textbf{55.5} \\
		\hline
	\end{tabular}}
    \vspace{1mm}
	\caption{Performance comparison with state-of-the-art methods on the Synthia$\rightarrow$Cityscapes task (mIoU: 16-class; mIoU*: 13-class). }
	\label{tab:synthia2city}
\end{table*}

\section{Experiments}
\subsection{Datasets and Implementation Details}
We follow the evaluation settings used in \cite{CAG}, and evaluate our proposed method with the source domain datasets \textbf{GTA5}~\cite{gta5} and \textbf{Synthia}\cite{synthia}, and the target domain dataset  \textbf{Cityscapes}~\cite{cityscapes}. The \textbf{GTA5} dataset shares 19 common categories with the \textbf{Cityscapes} dataset and all the irrelevant categories are ignored during training; the \textbf{Synthia} dataset shares 16 common categories with the \textbf{Cityscapes} dataset. Some previous works only train and test on a 13-category subset of the \textbf{Synthia} dataset, or train two models on both subset and the whole set for better performance. Here we follow the practice in \cite{intra_domain, fgan} to train a model only on the whole set and test it on both settings.

%The \textbf{Cityscapes} dataset is the target domain with $2,957$ of size $2048\times1024$ training images and $500$ validation images of the same resolution. \textbf{Cityscapes} has 19 categories in total. The \textbf{GTA5} and \textbf{Synthia} are two source domain datasets of computer generated synthetic images, which contain $24,966$ of size $1914\times1052$ training images and $9400$ of size $1280\times760$ training images respectively. 

According to Figure~\ref{fig:pipeline}, we first use the photometrically aligned source domain images to train an initial segmentation model $T_0$ in the coarse alignment step. Then, the model is trained in an iterative self-supervision manner with $K=6$ and $U=20k$, and the total number of training iterations is $140k$ which is comparable to all previous works~\cite{CAG,stuff_things}. In our experiments, $P_h=0.9$ and $p=10$ for the pseudo-labels (as in ~\cite{BDL}), and the regularization term $\beta$ in (\ref{equ:histgamma}) is $0.01$. The margin $\alpha$ is $0.2$ for the triplet loss. We use the standard color-jittering as the stochastic function $\tau(.)$ in both source and target domains as in \cite{fgan}. Following the same experimental settings in CAG~\cite{CAG}, we adopt DeepLab V3+(Resnet101)~\cite{deeplab}\footnote{${\rm https://github.com/RogerZhangzz/CAG\_UDA/issues/6}$} as our segmentation model. Our proposed method has been implemented in PyTorch~\cite{pytorch}, and all experiments are conducted on 4 NVIDIA GeForce 2080Ti GPUs with 1 sample on each GPU.  In the coarse alignment step, the stochastic gradient descent is used with momentum of $0.9$ and weight decay of $1e−4$. The learning rate is initially set to $5e−4$ and is decreased using the polynomial learning rate policy with power of $0.9$. The setting for the following iterative finetuning steps are exactly the same except we halve the learning rate to $2.5e−4$ to fine-tune previously trained models.
\subsection{Comparison with State-of-the-Art Methods}
In this section, we compare our method against all the existing state-of-the-art methods~\cite{intra_domain,stuff_things,CAG,FDA,fgan,BDL}, on both GTA5$\rightarrow$Cityscapes and Synthia$\rightarrow$Cityscapes tasks. 

For the GTA5$\rightarrow$Cityscapes task, according to Table~\ref{tab:gta2city}, it is clear that our proposed method outperforms all previous methods, achieving a new state-of-the-art mIoU at $56.1\%$ which is $5.9\%$ higher than previous state-of-the-art methods~\cite{CAG}. In general, our method achieves the best performance in many important categories, including `road', `sidewalk', `building', `light', `sky', `car', `person', `train', `motor', and `bike'. In particular, our model delivers a very good classification performance over `road', `sidewalk', `motor' and `bike' although some of these categories share very similar local appearances. This is because our category-oriented triplet loss focuses on the most confusing samples in different classes, and improves the generalization capability of the model. Moreover, the target consistency regularization in the target domain improves the classification accuracy of categories with a large intra-category variance, such as `building' and `sky'. 
%However, our model has relatively poor performance on `walls' and `fence'. In the target domain, the classification accuracy of such categories relies heavily on accurate pseudo-labels and category centers, which might not be available because the high-level features of such categories are too close to separate as they looks almost the same in appearances.
%and the proposed categorical data augmentation interferes the generation of the high-level features and make them even more indistinguishable.
% Although the category-oriented triplet loss aims to separate similar categories, it is only applied to the source domain.

The performance of the proposed method on Synthia$\rightarrow$Cityscapes is shown in Table~\ref{tab:synthia2city}. The Synthia dataset has a larger domain shift caused by perspective and layout in addition to photometric differences in comparison to the GTA5 dataset. But the overall performance of our model across all categories still surpasses the performance of other state-of-the-art methods, which demonstrates the effectiveness of our proposed techniques. 

In comparison to CAG~\cite{CAG} using the same segmentation model, our proposed modules achieve a significant performance improvement, which is $5.9\%$ in the GTA5$\rightarrow$Cityscapes task and $3.7\%$ in the Synthia$\rightarrow$Cityscapes task. We further show some of the segmentation samples in Figure~\ref{fig:qualitative} to qualitatively demonstrate the superiority of our method. Please refer to the supplementary document for \underline{\textbf{more qualitative examples}}.
%And a similar conclusion can be reached. That is, by emphasizing the inter-category differences and enhancing the intra-category variations in the source domain, our model achieves better classification accuracy on categories such as walls and fences. But our algorithm does not perform well for categories close to each other on the pixel locations such as bikes and riders. This might be alleviated by carefully tunning the intensity of the categorical data augmentation based on the characteristics of the pixel, which we leave as an open question in this paper.
%\subsection{Qualitative Studies}
%In addition to the high-quality results illustrated in Figure~\ref{tab:alignment} by our proposed global photometric alignment, we also have satisfactory segmentation results as shown in \ref{}

\begin{figure*}[ht]
	\centering
	\includegraphics[width=0.7\linewidth]{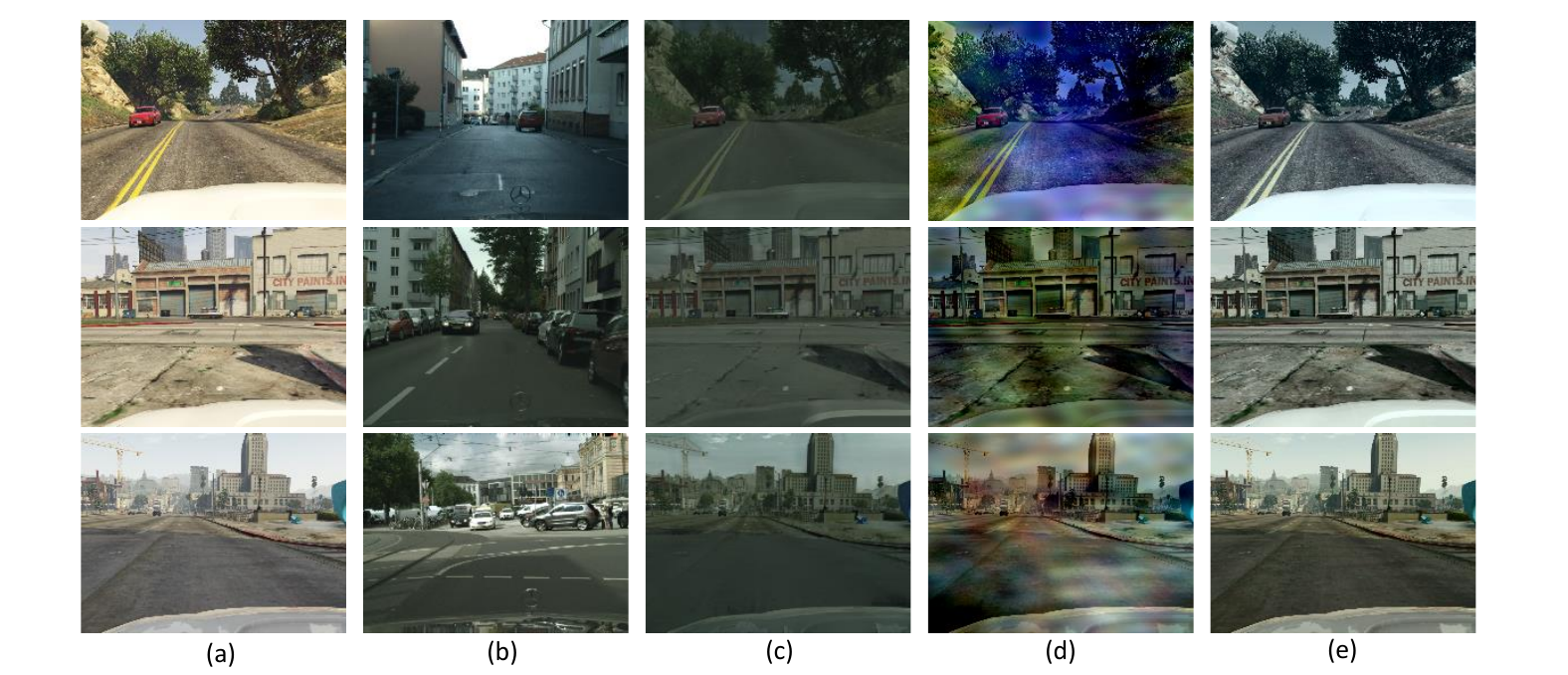} % Reduce the figure size so that it is slightly narrower than the column.`
	\caption{Qualitative analysis on global photometric alignment. (a) Input images, (b) reference image, (c) BDL-GAN\cite{BDL}, (d) Fourier Adaptation\cite{FDA}, (e) Global photometric alignment.}
	\label{fig:qualitative_gpa}
\end{figure*}

\subsection{Ablation Studies}
\textbf{Component Analysis.} Most previous works \cite{CAG,intra_domain,fgan,stuff_things} require a segmentation model pre-trained on the original source domain training set only, and we call this model the source-only model. Although we do not use the source-only model during training, we train one to provide a baseline to demonstrate that the primary performance gain comes from our proposed modules and pipeline. As shown in Table~\ref{tab:modules}, the performance of the source-only baseline using Deeplab v3+ is $37.6\%$, which is only slightly higher than that of the baseline using Deeplab v2 ($36.6\%$) reported in \cite{stuff_things,dagan1}, and our proposed pipeline improves the baseline performance by $18.5\%$. Following the same settings in previous state-of-the-art methods~\cite{CAG,intra_domain,fgan}, we further evaluate the impact of each proposed component on the performance of our model in the GTA5$\rightarrow$Cityscapes task by removing one component at a time. According to our experimental results,
the performance of the segmentation model has the most deterioration when the global photometric alignment module is removed. This is because in our coarse-to-fine pipeline, removing photometric alignment literally removes the first coarse alignment stage, and the resulting erroneous pseudo-labels are very detrimental to category-level feature distribution regularization. This also validates the necessity of a coarse alignment stage. Interestingly, although our target domain consistency regularization is simple, it has been proved to be very effective. This is because there are fewer training images in the target domain than the source domain, and filtering pseudo-labels with low confidence makes them even fewer. Our target domain consistency regularization increases the number of training samples in the target domain, therefore, giving rise to such a performance gain. The category-oriented triplet loss applied on the source domain also boosts the performance by $2.9\%$ as it exploits hard samples in the source domain.

\begin{table}[h]
	\centering
	\resizebox{0.85\linewidth}{!}{
	\begin{tabular}{l l l l l l}
		\hline
		& GPA & CTL & TCR & mIoU \\
		\hline
		Source only & & & & 37.6\\
		w/o GPA & & $\surd$ & $\surd$ & 47.5\\
		w/o CTL and TCR & $\surd$ & & & 47.3\\
		w/o CTL & $\surd$ & & $\surd$ & 53.2\\
		w/o TCR & $\surd$ & $\surd$ & & 53.1\\
		all & $\surd$ & $\surd$ & $\surd$ & 56.1\\
		\hline
	\end{tabular}}
    \vspace{1mm}
	\caption{Ablation study of the proposed components on the GTA5$\rightarrow$Cityscapes task. GPA: global photometric alignment, CTL: category-oriented triplet loss, TCR: target domain consistency regularization.}
	\label{tab:modules}
\end{table}

\textbf{Photometric Alignment.}
There are currently other methods, which can achieve the goal of coarse alignment, such as the GAN-based method in \cite{BDL, chen2017no} and the frequency-based method in \cite{FDA}. We substitute our proposed global photometric alignment with these two methods, and retrain our whole pipeline. The result is shown in Table~\ref{tab:ablations}. We also visualize some representative aligned images produced with different methods in Figure~\ref{fig:qualitative_gpa}. Our proposed GPA can generate the aligned image according to a randomly chosen target domain reference image, while the GAN-based model~\cite{BDL, chen2017no} performs deterministically and generates aligned images with a similar style, only covering part of the actual target domain image span. This explains why our proposed model works even better than the pre-trained deep adversarial model. Although the frequency-based method proposed in \cite{FDA} can generate style-transferred images randomly, the concatenation of frequencies usually introduces significant noises during training, which largely limits its final performance. 

Based on our observation, gamma correction on all three channels does not have sufficient alignment capability, while histogram matching on all three channels results in image artifacts. We have run a comparison for the coarse alignment stage and the result (Table~\ref{tab:ablations}) shows our hybrid scheme performs the best.

\textbf{Pseudo-labels.} In our proposed method, we only apply the category-oriented triplet loss to source domain category labels but not pseudo-labels in the target domain. Although target domain images with pseudo-labels can be used as supplementary samples when the pseudo-labels are of high confidence, our proposed triplet loss aims to deal with hard samples, and pseudo-labels of hard samples in the target domain are not reliable. In order to verify this, we include pseudo-labels in our category-oriented triplet loss, and the result is shown in Table~\ref{tab:ablations}. %Applying a careful selection strategy to the pseudo-labels in the target domain might improve the performance of our triplet loss, and will be our future work.

\begin{table}[h]
	\centering
	\resizebox{0.85\linewidth}{!}{
	\begin{tabular}{l l l}
		\hline
		Modules & Methods & mIoU\\
		\hline
		Image Align. & Frequency Align~\cite{FDA}. & 52.0\\
		& BDL-GAN~\cite{BDL} & 54.5\\
		& Photometric Align. & 56.1\\
		\hline
		GPA Scheme & Lab Gamma Correction & 44.5\\
		& Lab Histogram Match & 43.3\\
		& Hybrid & 47.3\\
		\hline
		Pseudo-labels & Triplet loss with pseudo-labels & 53.3\\
		& Triplet loss w/o pseudo-labels & 56.1\\
		\hline
	\end{tabular}}
    \vspace{1mm}
	\caption{Ablation studies of the image alignment plan, photometric alignment scheme, and using pseudo-labels for the category-oriented triplet loss on the GTA5$\rightarrow$Cityscapes task.}
	\label{tab:ablations}
\end{table}

\section{Conclusions}
In this paper, we propose a novel coarse-to-fine pipeline for domain adaptation semantic segmentation that smoothly integrates image-level alignment with category-level feature distribution regularization. In particular, we introduce a novel and efficient global photometric alignment module to coarsely align the source and target domains, and then, we propose a category-oriented triplet loss for the source domain and a target domain consistency regularization method to regularize the category-level feature distributions from a fine-grained category perspective. Experiments demonstrate that each of our proposed techniques improves the generalization capability of our model. And integrating them together results in a significant performance improvement in comparison to existing state-of-the-art unsupervised domain adapted semantic segmentation methods, demonstrating that solving image-level and category-level domain shifts simultaneously deserves more attention.

\section*{Acknowledgments}
This work was partially supported by National Key Research and Development Program of China (No.2020YFC2003902) and Hong Kong Research Grants Council through Research Impact Fund (Grant R-5001-18). H. Ma was supported by the Hong Kong PhD Fellowship.

{\small
\bibliographystyle{ieee_fullname}
\bibliography{egbib}

\begin{thebibliography}{10}\itemsep=-1pt

\bibitem{deeplab}
Liang-Chieh Chen, George Papandreou, Iasonas Kokkinos, Kevin Murphy, and Alan~L
  Yuille.
\newblock Deeplab: Semantic image segmentation with deep convolutional nets,
  atrous convolution, and fully connected crfs.
\newblock {\em IEEE transactions on pattern analysis and machine intelligence},
  40(4):834--848, 2017.

\bibitem{chen2017no}
Yi-Hsin Chen, Wei-Yu Chen, Yu-Ting Chen, Bo-Cheng Tsai, Yu-Chiang Frank~Wang,
  and Min Sun.
\newblock No more discrimination: Cross city adaptation of road scene
  segmenters.
\newblock In {\em Proceedings of the IEEE International Conference on Computer
  Vision}, pages 1992--2001, 2017.

\bibitem{cityscapes}
Marius Cordts, Mohamed Omran, Sebastian Ramos, Timo Rehfeld, Markus Enzweiler,
  Rodrigo Benenson, Uwe Franke, Stefan Roth, and Bernt Schiele.
\newblock The cityscapes dataset for semantic urban scene understanding.
\newblock In {\em Proceedings of the IEEE conference on computer vision and
  pattern recognition}, pages 3213--3223, 2016.

\bibitem{medical1}
Adrian~V Dalca, Evan Yu, Polina Golland, Bruce Fischl, Mert~R Sabuncu, and
  Juan~Eugenio Iglesias.
\newblock Unsupervised deep learning for bayesian brain mri segmentation.
\newblock In {\em International Conference on Medical Image Computing and
  Computer-Assisted Intervention}, pages 356--365. Springer, 2019.

\bibitem{autodrive2}
Di Feng, Christian Haase-Sch{\"u}tz, Lars Rosenbaum, Heinz Hertlein, Claudius
  Glaeser, Fabian Timm, Werner Wiesbeck, and Klaus Dietmayer.
\newblock Deep multi-modal object detection and semantic segmentation for
  autonomous driving: Datasets, methods, and challenges.
\newblock {\em IEEE Transactions on Intelligent Transportation Systems}, 2020.

\bibitem{DIP}
Rafael~C Gonzalez, Richard~Eugene Woods, and Steven~L Eddins.
\newblock {\em Digital image processing using MATLAB}.
\newblock Pearson Education India, 2004.

\bibitem{gan}
Ian Goodfellow, Jean Pouget-Abadie, Mehdi Mirza, Bing Xu, David Warde-Farley,
  Sherjil Ozair, Aaron Courville, and Yoshua Bengio.
\newblock Generative adversarial nets.
\newblock In {\em Advances in neural information processing systems}, pages
  2672--2680, 2014.

\bibitem{dagan5}
Judy Hoffman, Eric Tzeng, Taesung Park, Jun-Yan Zhu, Phillip Isola, Kate
  Saenko, Alexei Efros, and Trevor Darrell.
\newblock Cycada: Cycle-consistent adversarial domain adaptation.
\newblock In {\em International conference on machine learning}, pages
  1989--1998. PMLR, 2018.

\bibitem{fcns}
Judy Hoffman, Dequan Wang, Fisher Yu, and Trevor Darrell.
\newblock Fcns in the wild: Pixel-level adversarial and constraint-based
  adaptation.
\newblock {\em arXiv preprint arXiv:1612.02649}, 2016.

\bibitem{medical3}
Yuankai Huo, Zhoubing Xu, Yunxi Xiong, Katherine Aboud, Prasanna Parvathaneni,
  Shunxing Bao, Camilo Bermudez, Susan~M Resnick, Laurie~E Cutting, and
  Bennett~A Landman.
\newblock 3d whole brain segmentation using spatially localized atlas network
  tiles.
\newblock {\em NeuroImage}, 194:105--119, 2019.

\bibitem{sda1}
Naoto Inoue, Ryosuke Furuta, Toshihiko Yamasaki, and Kiyoharu Aizawa.
\newblock Cross-domain weakly-supervised object detection through progressive
  domain adaptation.
\newblock In {\em Proceedings of the IEEE conference on computer vision and
  pattern recognition}, pages 5001--5009, 2018.

\bibitem{da2}
Liming Jiang, Changxu Zhang, Mingyang Huang, Chunxiao Liu, Jianping Shi, and
  Chen~Change Loy.
\newblock Tsit: A simple and versatile framework for image-to-image
  translation.
\newblock {\em arXiv preprint arXiv:2007.12072}, 2020.

\bibitem{uda1}
Seunghyeon Kim, Jaehoon Choi, Taekyung Kim, and Changick Kim.
\newblock Self-training and adversarial background regularization for
  unsupervised domain adaptive one-stage object detection.
\newblock In {\em Proceedings of the IEEE International Conference on Computer
  Vision}, pages 6092--6101, 2019.

\bibitem{BDL}
Yunsheng Li, Lu Yuan, and Nuno Vasconcelos.
\newblock Bidirectional learning for domain adaptation of semantic
  segmentation.
\newblock In {\em Proceedings of the IEEE Conference on Computer Vision and
  Pattern Recognition}, pages 6936--6945, 2019.

\bibitem{oldfcn}
Jonathan Long, Evan Shelhamer, and Trevor Darrell.
\newblock Fully convolutional networks for semantic segmentation.
\newblock In {\em Proceedings of the IEEE conference on computer vision and
  pattern recognition}, pages 3431--3440, 2015.

\bibitem{take_a_look}
Yawei Luo, Liang Zheng, Tao Guan, Junqing Yu, and Yi Yang.
\newblock Taking a closer look at domain shift: Category-level adversaries for
  semantics consistent domain adaptation.
\newblock In {\em Proceedings of the IEEE Conference on Computer Vision and
  Pattern Recognition}, pages 2507--2516, 2019.

\bibitem{cgan}
Mehdi Mirza and Simon Osindero.
\newblock Conditional generative adversarial nets.
\newblock {\em arXiv preprint arXiv:1411.1784}, 2014.

\bibitem{intra_domain}
Fei Pan, Inkyu Shin, Francois Rameau, Seokju Lee, and In~So Kweon.
\newblock Unsupervised intra-domain adaptation for semantic segmentation
  through self-supervision.
\newblock In {\em Proceedings of the IEEE/CVF Conference on Computer Vision and
  Pattern Recognition}, pages 3764--3773, 2020.

\bibitem{pytorch}
Adam Paszke, Sam Gross, Soumith Chintala, Gregory Chanan, Edward Yang, Zachary
  DeVito, Zeming Lin, Alban Desmaison, Luca Antiga, and Adam Lerer.
\newblock Automatic differentiation in pytorch.
\newblock 2017.

\bibitem{sda2}
Jiangtao Peng, Weiwei Sun, Li Ma, and Qian Du.
\newblock Discriminative transfer joint matching for domain adaptation in
  hyperspectral image classification.
\newblock {\em IEEE Geoscience and Remote Sensing Letters}, 16(6):972--976,
  2019.

\bibitem{gta5}
Stephan~R Richter, Vibhav Vineet, Stefan Roth, and Vladlen Koltun.
\newblock Playing for data: Ground truth from computer games.
\newblock In {\em European conference on computer vision}, pages 102--118.
  Springer, 2016.

\bibitem{synthia}
German Ros, Laura Sellart, Joanna Materzynska, David Vazquez, and Antonio~M
  Lopez.
\newblock The synthia dataset: A large collection of synthetic images for
  semantic segmentation of urban scenes.
\newblock In {\em Proceedings of the IEEE conference on computer vision and
  pattern recognition}, pages 3234--3243, 2016.

\bibitem{consistency}
Mehdi Sajjadi, Mehran Javanmardi, and Tolga Tasdizen.
\newblock Regularization with stochastic transformations and perturbations for
  deep semi-supervised learning.
\newblock In {\em Advances in neural information processing systems}, pages
  1163--1171, 2016.

\bibitem{fixmatch}
Kihyuk Sohn, David Berthelot, Chun-Liang Li, Zizhao Zhang, Nicholas Carlini,
  Ekin~D Cubuk, Alex Kurakin, Han Zhang, and Colin Raffel.
\newblock Fixmatch: Simplifying semi-supervised learning with consistency and
  confidence.
\newblock {\em arXiv preprint arXiv:2001.07685}, 2020.

\bibitem{autodrive1}
Michael Treml, Jos{\'e} Arjona-Medina, Thomas Unterthiner, Rupesh Durgesh,
  Felix Friedmann, Peter Schuberth, Andreas Mayr, Martin Heusel, Markus
  Hofmarcher, Michael Widrich, et~al.
\newblock Speeding up semantic segmentation for autonomous driving.
\newblock In {\em MLITS, NIPS Workshop}, volume~2, 2016.

\bibitem{dagan1}
Yi-Hsuan Tsai, Wei-Chih Hung, Samuel Schulter, Kihyuk Sohn, Ming-Hsuan Yang,
  and Manmohan Chandraker.
\newblock Learning to adapt structured output space for semantic segmentation.
\newblock In {\em Proceedings of the IEEE Conference on Computer Vision and
  Pattern Recognition}, pages 7472--7481, 2018.

\bibitem{dagan4}
Tuan-Hung Vu, Himalaya Jain, Maxime Bucher, Matthieu Cord, and Patrick
  P{\'e}rez.
\newblock Advent: Adversarial entropy minimization for domain adaptation in
  semantic segmentation.
\newblock In {\em Proceedings of the IEEE conference on computer vision and
  pattern recognition}, pages 2517--2526, 2019.

\bibitem{medical2}
Guotai Wang, Wenqi Li, Tom Vercauteren, and Sebastien Ourselin.
\newblock Automatic brain tumor segmentation based on cascaded convolutional
  neural networks with uncertainty estimation.
\newblock {\em Frontiers in computational neuroscience}, 13:56, 2019.

\bibitem{fgan}
Haoran Wang, Tong Shen, Wei Zhang, Lingyu Duan, and Tao Mei.
\newblock Classes matter: A fine-grained adversarial approach to cross-domain
  semantic segmentation.
\newblock {\em arXiv preprint arXiv:2007.09222}, 2020.

\bibitem{stuff_things}
Zhonghao Wang, Mo Yu, Yunchao Wei, Rogerio Feris, Jinjun Xiong, Wen-mei Hwu,
  Thomas~S Huang, and Honghui Shi.
\newblock Differential treatment for stuff and things: A simple unsupervised
  domain adaptation method for semantic segmentation.
\newblock In {\em Proceedings of the IEEE/CVF Conference on Computer Vision and
  Pattern Recognition}, pages 12635--12644, 2020.

\bibitem{dagan3}
Zuxuan Wu, Xintong Han, Yen-Liang Lin, Mustafa Gokhan~Uzunbas, Tom Goldstein,
  Ser Nam~Lim, and Larry~S Davis.
\newblock Dcan: Dual channel-wise alignment networks for unsupervised scene
  adaptation.
\newblock In {\em Proceedings of the European Conference on Computer Vision
  (ECCV)}, pages 518--534, 2018.

\bibitem{dagan2}
Shaoan Xie, Zibin Zheng, Liang Chen, and Chuan Chen.
\newblock Learning semantic representations for unsupervised domain adaptation.
\newblock In {\em International Conference on Machine Learning}, pages
  5423--5432, 2018.

\bibitem{FDA}
Yanchao Yang and Stefano Soatto.
\newblock Fda: Fourier domain adaptation for semantic segmentation.
\newblock In {\em Proceedings of the IEEE/CVF Conference on Computer Vision and
  Pattern Recognition}, pages 4085--4095, 2020.

\bibitem{CAG}
Qiming Zhang, Jing Zhang, Wei Liu, and Dacheng Tao.
\newblock Category anchor-guided unsupervised domain adaptation for semantic
  segmentation.
\newblock In {\em Advances in Neural Information Processing Systems}, pages
  435--445, 2019.

\bibitem{da1}
Yabin Zhang, Bin Deng, Kui Jia, and Lei Zhang.
\newblock Label propagation with augmented anchors: A simple semi-supervised
  learning baseline for unsupervised domain adaptation.
\newblock {\em arXiv preprint arXiv:2007.07695}, 2020.

\bibitem{crf}
Shuai Zheng, Sadeep Jayasumana, Bernardino Romera-Paredes, Vibhav Vineet,
  Zhizhong Su, Dalong Du, Chang Huang, and Philip~HS Torr.
\newblock Conditional random fields as recurrent neural networks.
\newblock In {\em Proceedings of the IEEE international conference on computer
  vision}, pages 1529--1537, 2015.

\bibitem{cgan2}
Jun-Yan Zhu, Taesung Park, Phillip Isola, and Alexei~A Efros.
\newblock Unpaired image-to-image translation using cycle-consistent
  adversarial networks.
\newblock In {\em Proceedings of the IEEE international conference on computer
  vision}, pages 2223--2232, 2017.

\end{thebibliography}
}

\end{document}